\documentclass[letterpaper, 10 pt, conference]{ieeeconf} 
\IEEEoverridecommandlockouts  
\usepackage{subfiles}
\usepackage{float}
\usepackage{graphicx}
\usepackage[colorlinks]{hyperref}
\usepackage{gensymb}

\title{\LARGE \bf
GODSAC*: Graph Optimized DSAC* for Robot Relocalization
}

\author{Alphonsus Adu-Bredu$^{1*}$ \hspace{0.5cm} Noah Del Coro$^{1*}$ \hspace{0.5cm} Tianyi Liu$^{1*}$       
\thanks{$^{1}$Alphonsus Adu-Bredu, Noah Del Coro, Tianyi Liu are with the Robotics Institute and Department of Electrical Engineering and Computer Science, University of Michigan, Ann Arbor, MI, USA.
        {\tt\small [adubredu|ndelcoro|tannertl]@umich.edu}}
\thanks{$^{*}$Equal Contribution}%
}

\begin{document}
	
\maketitle
\thispagestyle{empty}
\pagestyle{empty}

\begin{abstract} 
Deep learning based camera pose estimation from monocular camera images has seen a recent uptake in Visual SLAM research. Even though such pose estimation approaches have excellent results in small confined areas like offices and apartment buildings, they tend to do poorly when applied to larger areas like outdoor settings, mainly because of the scarcity of distinctive features. We propose GODSAC* as a camera pose estimation approach that augments pose predictions from a trained neural network with noisy odometry data through the optimization of a pose graph. GODSAC* outperforms the state-of-the-art approaches in pose estimation accuracy, as we demonstrate in our experiments. Our open-source implementation of GODSAC* and a summary video can be found at this url: \href{https://github.com/alphonsusadubredu/godsacstar}{https://github.com/alphonsusadubredu/godsacstar}
\end{abstract}
\begin{figure*}
    \centering
    \includegraphics[width=\textwidth]{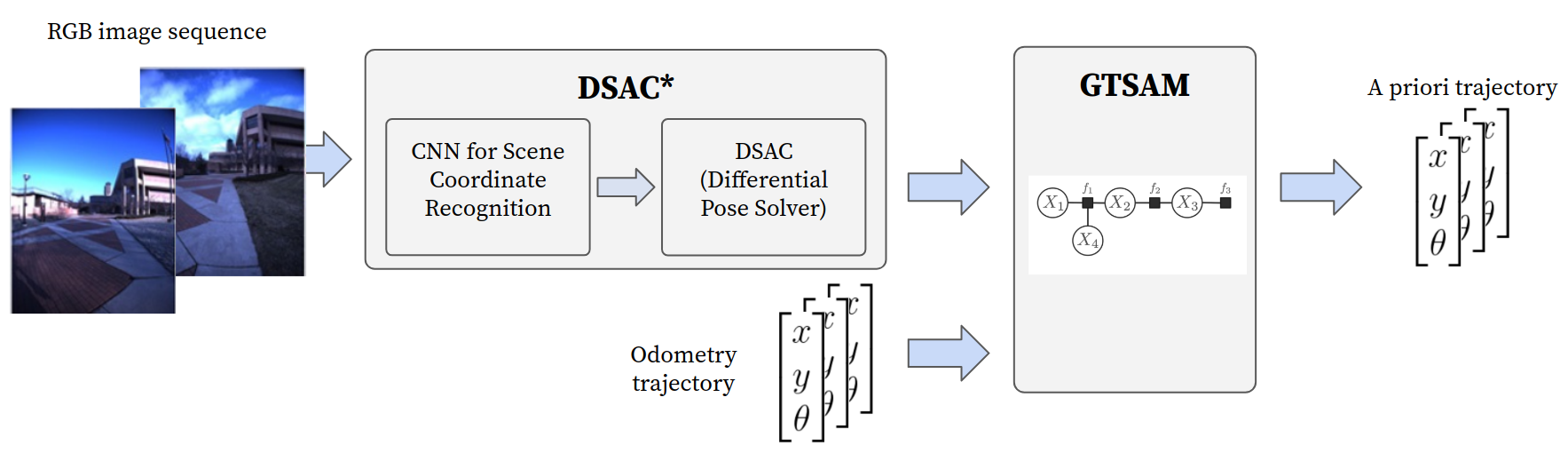}
    \caption{An Overview of the GODSAC* framework. Given an RGB image sequence, DSAC* outputs a prediction of the camera pose. This prediction is augmented with the corresponding odometry sequence into a pose graph. The pose graph is optimized with GTSAM's implementation of iSAM2 and outputs the camera's estimated trajectory.}
    \label{fig:overview}
\end{figure*}
\section{Introduction} 
The recent advancement in Deep Learning research has opened the door to numerous applications, especially in the field of computer vision research. Prominent amongst these applications is the task of estimating the pose of a monocular camera from its RGB images only. Over the years, monocular cameras have grown cheaper  and can be found in almost every mobile device. The potential advantages for using such ubiquitous devices for pose estimation and Simultaneous Localization and Mapping (SLAM) are enormous. 

Numerous works have focused on the usage of monocular cameras for indoor camera pose estimation and localization \cite{ababsa2004robust, bleser2006online,camposeco2018hybrid, en2018rpnet, kendall2015posenet,kendall2017geometric,melekhov2017relative, ohayon2006robust,quan1999linear, rambach2016learning,rodrigues2010camera}. Amongst these, works such as \cite{kendall2015posenet,kendall2017geometric, camposeco2018hybrid, en2018rpnet,melekhov2017relative,rambach2016learning} have focused on the use of convolutional neural networks to predict the camera's pose given an RGB image. Although these approaches tend to have excellent results in indoor settings where there is an abundance of distinct features, their excellent performance do not extend to large outdoor areas where features may be scarce and where there is a potential for significant amount of drift in the pose estimation results.

To tackle this problem, we propose Graph Optimized DSAC* (GODSAC*) as an approach for performing outdoor monocular camera pose estimation from RGB images only. GODSAC* augments camera pose predictions from a trained DSAC* neural network \cite{dsacstar} with noisy odometry data through the iterative optimization of a pose graph using the incremental smoothing and mapping framework, iSAM2 \cite{isam2}. Jointly optimizing the pose predictions from DSAC* with the noisy odometry data produces significantly better results for large outdoor scenes, as we demonstrate in our experiments. Figure \ref{fig:overview} depicts an overview of our approach. 

We compare GODSAC* with state-of-the-art methods in  camera pose prediction in our experiments. GODSAC* outperforms these approaches in terms of translation and rotation errors, across all the datasets we use in our experiments. 

In the following sections, we first talk briefly about related work and describe our approach in detail. Next, we describe our experimental setup, showcase both qualitative and quantitative results and discuss the implications of our experimental results.
\section{Related Work}
The problem of estimating camera poses from monocular RGB images has been around for a while. Early approaches for solving this problem \cite{orbslam, bundle, monoslam,invmonoslam, sfmcaus} used either filtering \cite{monoslam,invmonoslam,sfmcaus} or feature-based \cite{orbslam} approaches that used hand-designed feature extractors to extract certain features in the input RGB images which are tracked over time to estimate the pose of the camera. In MonoSLAM \cite{monoslam, invmonoslam}, every frame of incoming RGB images is processed by a filter to jointly estimate feature locations and the camera pose. The drawback of these approaches is that, they are slow, computationally expensive, and accumulate linearization errors. Feature-based approaches \cite{orbslam}, on the other hand, estimate the camera pose on only select frames and features, making them more computationally efficient for indoor scenes, which usually have rich and distinctive features. These approaches cannot extend to large outdoor environments because of the scarcity of distinctive features.

With the latest advancements in Deep Learning research on computer vision problems like object detection, image segmentation and classification and image captioning, researchers have recently focused their attention to solving the problem of camera pose estimation using deep learning. The benefits of choosing a deep learning based approach over the classical feature-based methods are 1) faster inference times of deep learning approaches (milliseconds vs minutes), 2) lower memory requirements of deep learning based approaches (megabytes vs gigabytes), 3) absence of feature engineering.   PoseNet \cite{kendall2015posenet} is the first deep learning-based approach that proposed the use of trained neural network to regress the absolute camera pose given an image. PoseNet has a modified truncated GoogLeNet\cite{googlenet} architecture with its softmax replaced with  a sequence of fully connected layers to output the absolute pose of a camera when given an image. The drawbacks of PoseNet however were that, 1) it overfits to the training set and does not generalize well to new scenes and 2) its estimation error in indoor and outdoor scenes was much larger than that of feature-based approaches. DSAC* \cite{dsacstar} is a neural network architecture based on the ResNet architecture \cite{resnet} that performs more accurate absolute pose estimation than PoseNet. In this work, we use DSAC* to perform the initial pose estimation from the input RGB images.

To operate successfully, mobile robots need to estimate their pose without the aid of external referencing systems like GPS. SLAM approaches have historically been tackled by two main classes of approaches namely filtering and smoothing. Filtering approaches \cite{kf1,kf2,pf1,pf2,pf3} model the SLAM problem as an online state estimation where the state is the current position of the robot and the map. Estimation is updated through a bayesian update process as new measurements are taken by the robot. Filtering approaches are often referred to as Online SLAM methods because of their incremental nature. These approaches are strictly Markovian by design and only consider the most recent measurement during inference. As a result, estimates are often noisy and have significant drift. The other class of SLAM approaches, smoothing \cite{smoothing}, formulate the SLAM problem into a graph where nodes represent the robot poses or landmarks and edges between two nodes represent sensor measurements that constrain the connected poses. After constructing this graph, the poses are jointly inferred by solving a large least-squares error minimization problem that solves for pose values that are consistent with the measurements. Since this optimization is performed over the entire graph, the smoothing approach to SLAM return significantly more accurate pose estimates and maps. 

GTSAM \cite{gtsam} is a C++ library that implements iSAM2 \cite{isam2}, an incremental smoothing and mapping algorithm for graph optimization. In this work, we augment the pose estimation from DSAC* with noisy odometry into a pose graph which we jointly optimize using the iSAM2 algorithm. This optimization returns significantly more accurate pose estimations when run on difficult outdoor scenes datasets than current state-of-the-art methods like PoseNet++ \cite{posenetplus}.

Complex robotics systems often require some form of communication protocol to enable data transfer across submodules of the system. Two of the main open-source communication protocols employed in modern robots are Robot Operating System (ROS) \cite{ros} and Lightweight Communications and Marshalling (LCM) \cite{lcm}. We use ROS as the main communication protocol in this work because it offers better tools for sending and receiving high density data like images and pointclouds at high frequencies. It also has a much larger community usage in both industry and academia than LCM. 
\section{Methodology}

\begin{figure}
    \centering
    \includegraphics[width=1\columnwidth]{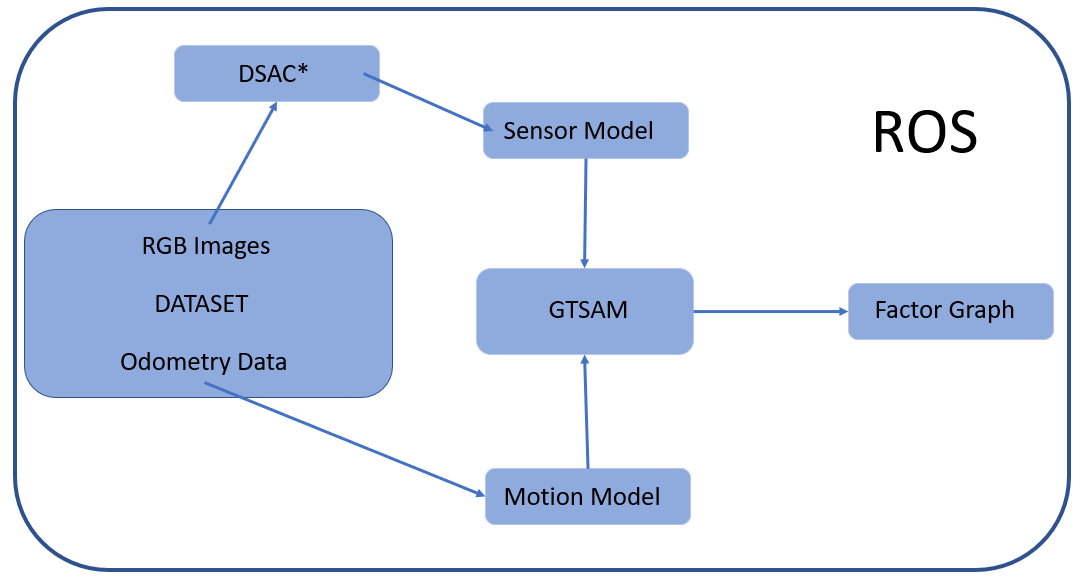}
    \caption{Pipeline for GODSAC*}
    \label{fig:pipeline}
\end{figure}

The pipeline of GODSAC* can be divided into two stages: the neural network-based pose estimator stage (DSAC* \cite{dsacstar}) and the GTSAM-based iSAM2 optimizer\cite{gtsam} stage. The DSAC* stage needs to be trained using the provided RGB images of the scene and ground truth poses corresponding to each image. The trained model can then inference and predict a camera pose given a RGB image. We then add the measured pose and corresponding odometry information into the iSAM2 optimizer to conduct the back-end optimization (GTSAM stage). As the pose measurements accumulate, an optimized trajectory with higher accuracy is obtained by the optimizer. Since there are several nodes in the localization process, to simplify the development and run time process, we utilize the ROS to serve as the communication architecture, which can be visualized in Figure \ref{fig:pipeline}.

\subsection{DSAC* Front-end} 
DSAC* \cite{dsacstar} is a versatile, neural network framework for visual camera re-localization based on geometry feature. It consists of two parts — (1) scene coordinate regression to predict the corresponding 3D scene point for a given 2D pixel location, and (2) Random Sample Consensus (RANSAC) based estimation to obtain the camera pose from the predicted scene coordinates.


DSAC* is based on ResNet \cite{resnet}, which outperformed its predecessor networks such as VGG \cite{vgg} or GoogleNet \cite{googlenet} network used in PoseNet \cite{kendall2015posenet} and many other pose estimation frameworks. The authors of DSAC* implement it using the Pytorch with some C++ libraries to compute the gradients for RANSAC. With the computed gradients of RANSAC, we can utilize the auto-grad feature of Pytorch to easily train the neural network. With the support of these cutting-edge frameworks, it is expected to predict both the scene coordinates and camera pose with considerably high accuracy.

\subsection{GTSAM Back-end}
In the back-end stage, we solve the problem of localizing the robot using the factor graph approach. Unlike normal factor graph problem, we add an additional factor of the pose estimation from the DSAC* and connect it to the robot state.

To realize the factor graph optimization, we utilize the iSAM2 optimizer as it has robust smoothing and mapping performance. As shown in Figure \ref{fig:overview}, the optimizer needs to handle input pose measurement from DSAC* and odometry data. We use the C++ language to implement the iSAM2 optimizer as a ROS service, which receives the poses and odometry information from other nodes. We choose C++ as the programming language because it has the lower latency than python, in which previous re-localization implementations have been based.

The poses generated by DSAC* are 6-DOF, but we are targeting ground vehicle locomotion — so we simplify the SE(3) pose into an SE(2) pose by assuming no translation in vehicle Z-axis and no rotation in roll and pitch axis. The frequency of the image sequence is much lower than the odometry data, so we accumulate the odometry data between two consequent images and treat the accumulated result as one edge in the factor graph. For the pose estimation factor, we use the estimated pose from DSAC* and a prior covariance matrix from the training stage because we assume the user scenario of the training and predicting does not vary much.

\subsection{ROS-based Communication Architecture}
ROS is a widely-used open-source package which provides libraries and tools to strengthen robot applications. ROS could serve as a communication interface among multiple nodes and services, and have native support for C++ and Python. Since the DSAC* pose estimation is based on Python and Pytorch, while the iSAM2 optimizer we implement is based on C++, it is natural that we utilize ROS, which is language agnostic. Furthermore, we use ROS to visualize the optimized result with the RVIZ package. To simulate a real application, we record the image sequence into a rosbag so the user can visualize the real-time optimized result.


\begin{table*}
\centering
\caption{Front-end Results on King's College Dataset and NCLT Dataset}
 \begin{tabular}{||c | c c | c | c c||} 
 \hline
 Dataset & Train Frames & Test Frames & Spatial Extent & PoseNet Mean Error & DSAC* Mean Error \\ [0.5ex] 
 \hline
 King's College & 1220 & 343 & 140m x 40m & 4.98m, 3.52\degree & 0.173m, 0.3\degree\\ 
 NCLT & 4500 & 1000 & 300m x 150m & 31.3m, 16.3\degree & 12.3m, 2.5\degree\\ 
 \hline
 \end{tabular}
 \label{tab:front_estimation}
\end{table*}

\section{Experiments}
The performance of our proposed method is evaluated using the University of Michigan North Campus Long-Term Vision and LIDAR (NCLT) dataset \cite{nclt} and King's College scene from the Cambridge Landmark dataset \cite{kendall2015posenet}. For the NCLT dataset, we utilized its RGB images and odometry data. For the Cambridge dataset, we only used its RGB images, then simulated noisy odometry from ground truth poses, since the dataset was not recorded with an onboard odometry sensor. The full GODSAC* pipeline is implemented as a ROS package with a Python script for DSAC* inferencing and a C++ node for GTSAM iSAM2 optimization.
\subsection{Datasets}
We conduct our experiments on the following datasets to evaluate the performance of the GODSAC* under different scenarios. The choices of datasets in our work are very similar to PoseNet++ as we try to validate the improvement of our solution based on their results in \cite{posenetplus}.
\subsubsection{King's College}
\begin{figure}
    \centering
    \includegraphics[width=1\columnwidth]{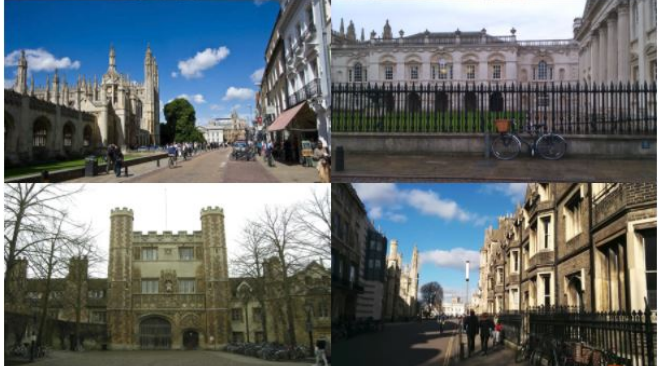}
    \caption{Sample Images of the Cambridge Landmarks Dataset}
    \label{fig:cambridge sample}
\end{figure}

King's College is a scene of the Cambridge Landmark dataset, which has been widely used to verify the performance of the pose estimation solution since it was first introduced in the PoseNet paper \cite{kendall2015posenet}. The objective to use this dataset is that we want to compare the pose estimation results of DSAC* \cite{dsacstar} and the PoseNet \cite{kendall2015posenet}. Sample images from this dataset are shown in Figure \ref{fig:cambridge sample}.
\subsubsection{NCLT}
\begin{figure}
    \centering
    \includegraphics[width=1\columnwidth]{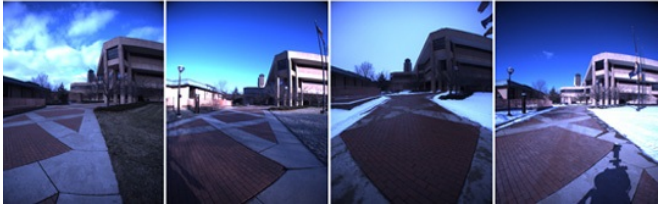}
    \caption{Sample Images of the NCLT Dataset}
    \label{fig:nclt sample}
\end{figure}

The NCLT dataset \cite{nclt} is collected on the north campus of the University of Michigan. It is famous for its long-term capturing of the same environment for 15 months across various weather and lightning conditions. There are 27 sessions in the NCLT dataset. For each session, the duration is approximately 1.29 hours and the distance travelled is around 5.5km. Furthermore, the NCLT dataset has provided the odometry data, which is ideal for our use case to conduct the backend optimization. We chose NCLT because it provides all the sensor data required for our proposed solution, its large environment is a challenge compared to other datasets, and it allowed us to directly compare with the results of PoseNet++ \cite{posenetplus}. Sample images from this dataset are shown in Figure \ref{fig:nclt sample}. 
\subsubsection{Comparison between King's College and NCLT}
According to the Figure \ref{fig:cambridge sample}, we find that in the King's college dataset, there is a large portion of the building information in each image. Besides, all the captured buildings tend to have many geometry features, which have been proved useful and important in geometry-based pose estimation solution such as DSAC* \cite{poseintroduction, dsacstar}. However, images shown in Figure \ref{fig:nclt sample} suggests that the images in the NCLT dataset tend to contain fewer geometric features because a large portion of the image is covered by the road, sky etc. Besides, the lighting conditions and weather might vary in different sessions of NCLT, which challenge the capability of the neural network to learn the latent geometric features.
\subsection{Front-end Training}
DSAC* supports both RGB and RGB-D images. In our case, the King's college dataset and the NCLT dataset have only RGB images, so we train the DSAC* in the RGB mode. In general, the DSAC* training can be divided into the initial stage and the end-to-end stage. In the initial stage, we train the network to predict the scene coordinates based on a pre-defined target depth of 10 meters (default value in DSAC*). In the end-to-end stage, we re-train the network to fine-tune the parameters with a loss function of the error between the pose estimation after RANSAC solver and the ground truth pose.

\begin{figure}
    \centering
    \includegraphics[width=1\columnwidth]{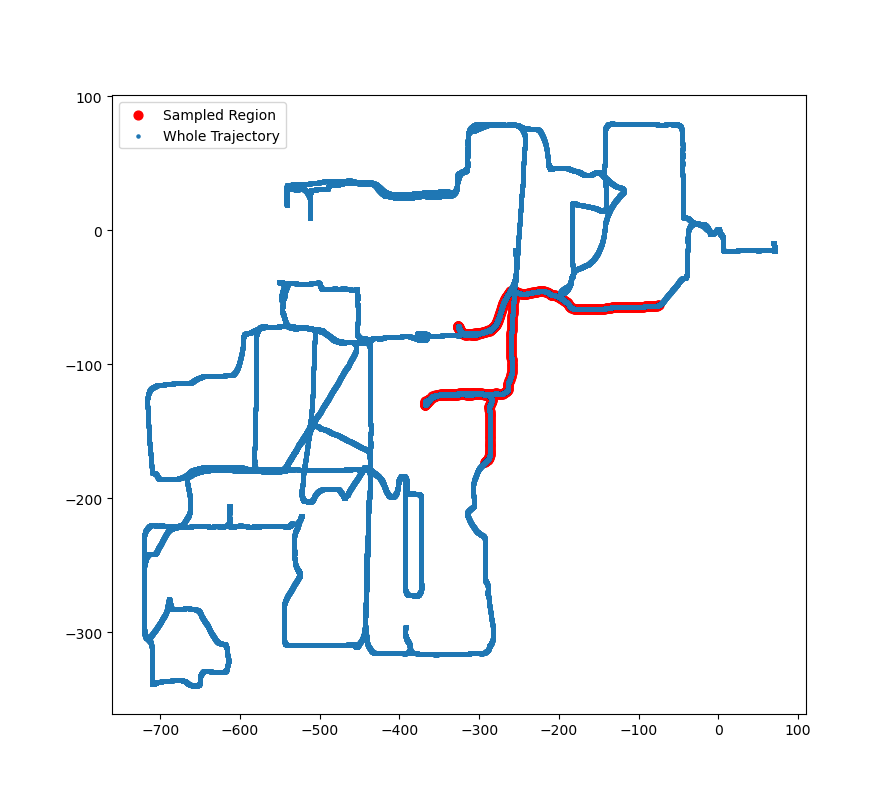}
    \caption{Selected Experiment Region versus Whole Trajectory}
    \label{fig:sample trajectory}
\end{figure}

For the NCLT dataset, we sampled from the same region as the authors in PoseNet++ did for direct benchmarking. The sampled region versus the whole trajectory is shown in Figure \ref{fig:sample trajectory}. Though the sampled region is a small subset of the whole trajectory, it is still 8 times larger than the King's College. A comparison between parameters and results of the two datasets can be seen in Table \ref{tab:front_estimation}. A pre-trained network model was used to evaluate DSAC* performance on King's College.

As to the training set and test set, the sequences we choose for training are: 2012-01-08, 2012-03-17 and 2012-10-28. For each sequence, we sample only 1500 images for training. According to our inspection of the training data, we consider the NCLT to be a challenging dataset for DSAC*, as it contains few extractable features. For the test set, we choose the 2012-01-08 sequence and sample another 1000 images that are distinct from the training set.

\begin{figure}
    \centering
    \includegraphics[width=1\columnwidth]{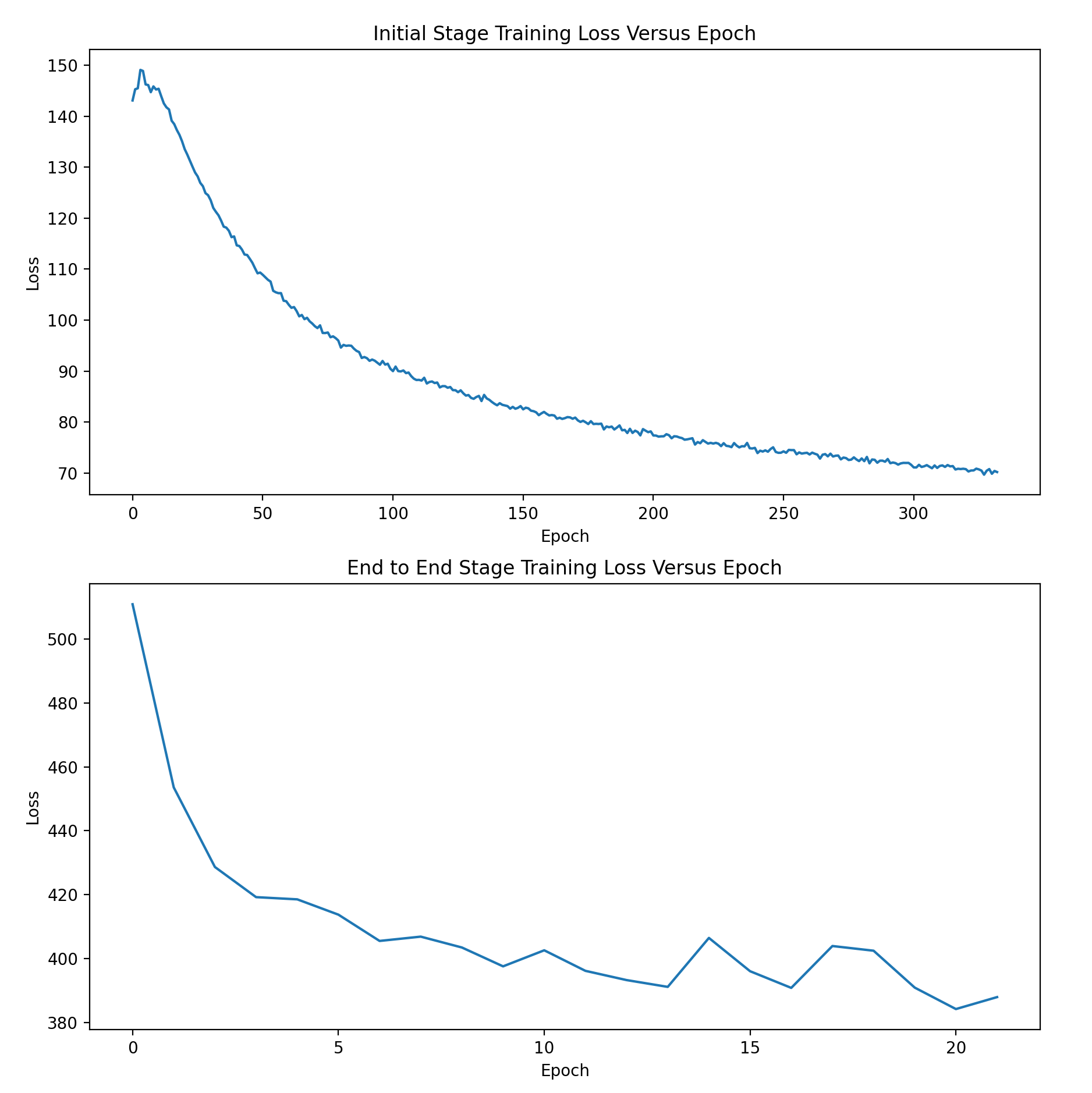}
    \caption{Loss of Two Stages in Training DSAC* on NCLT Dataset}
    \label{fig:loss}
\end{figure}

We plot the loss in the initial stage and end-to-end stage for the NCLT dataset training in Figure \ref{fig:loss}. The loss in both stages decreases optimally. We ran the initial stage for 1,500,000 iterations and end-to-end stage for 100,000 iterations. The whole process of training took around 20 hours on an Ubuntu PC with a Nvidia GTX 1080 GPU.

\subsection{Front-end Testing}
\begin{figure}
    \centering
    \includegraphics[width=1\columnwidth]{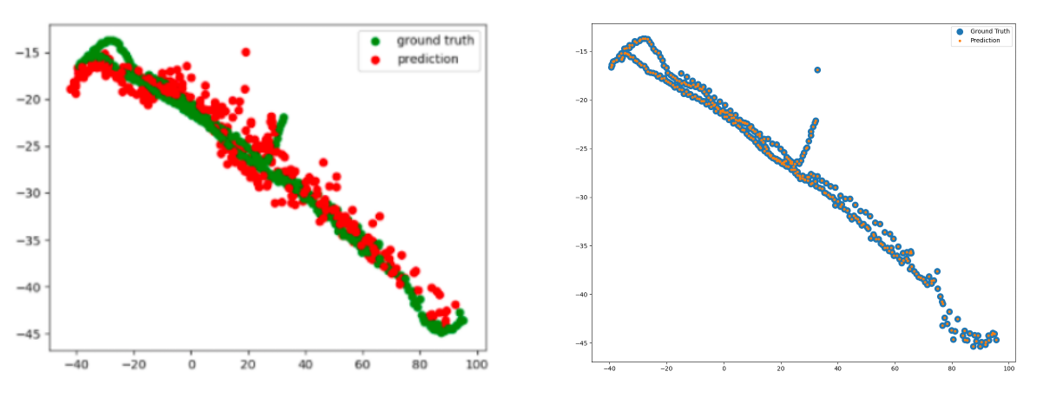}
    \caption{Pose Estimation Results on King's College Dataset for (i) PoseNet and (ii) DSAC*}
    \label{fig:front_king_compare}
\end{figure}

\begin{figure}
    \centering
    \includegraphics[width=1\columnwidth]{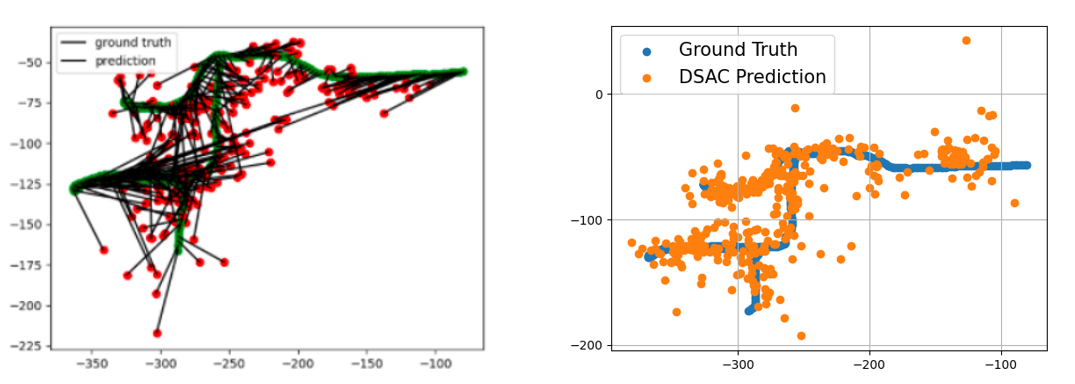}
    \caption{Pose Estimation Result on NCLT Dataset for (i) PoseNet and (ii) DSAC*}
    \label{fig:front_nclt_compare}
\end{figure}

We tested DSAC* on the King's College dataset and the NCLT dataset. We also compare the results with the PoseNet results reported in \cite{posenetplus}. The corresponding plots are shown in Figures \ref{fig:front_king_compare} and \ref{fig:front_nclt_compare}. We list the quantitative comparison in Table \ref{tab:front_estimation}, from which we can may find the DSAC* outperforms PoseNet significantly in terms of the translation and rotation errors. According to the result, we find that the DSAC* could generate very accurate pose estimation result at the environment where the geometric features are obvious e.g., King's college dataset. For the environment where the geometric features are hard to detect and the spatial coverage is very large, DSAC* tends to introduce larger errors even though it performs much better than PoseNet. So we seek to improve the pose estimation performance with DSAC* by back-end optimization, which is discussed in the following section.

\subsection{Back-end Optimization}

\begin{table}[H]
\centering
\caption{Covariance used in GTSAM}
 \begin{tabular}{||c | c c c ||} 
 \hline
 Covariance & x (m) & y (m) & $\theta$ (radian) \\ [0.5ex] 
 \hline
 Odometry & $0.024$ & $0.021$ & $0.056$\\ 
 Measurement & $15.621$ & $10.359$ & $0.086$ \\ 
 \hline
 \end{tabular}
 \label{tab:GTSAMCov}
\end{table}

\begin{figure}
    \centering
    \includegraphics[width=1\columnwidth]{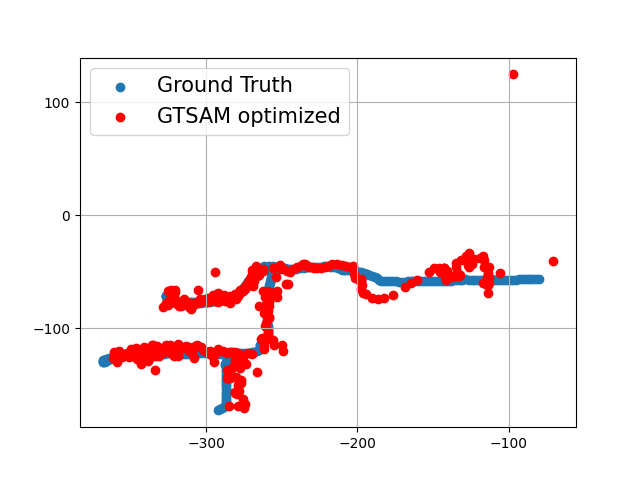}
    \caption{GTSAM optimized DSAC* trajectory}
    \label{fig:gtsam_NCLT}
\end{figure}

The covariance values used in experiments can be seen in Table \ref{tab:GTSAMCov}. The odometry covariance was determined as the mean in the NCLT covariance data. The measurement covariance were determined by calculating the covariance over all DSAC* NCLT test results. The optimized trajectory over our NCLT test set can be seen in Figure \ref{fig:gtsam_NCLT}. The trajectory begins from coordinate [-120, -50] on the right of the figure. A bad prior vertex was given to GTSAM to showcase its capability to correct the localization estimation. As seen in Figure \ref{fig:gtsam_NCLT}, the optimized trajectory quickly converged to ground truth after a few iterations. 

As shown in Table \ref{tab:gtsam_err}, the GTSAM further reduced the root mean squared error in DSAC* inference results.
\begin{table}[H]
\centering
\caption{DSAC* Inference VS GTSAM optimization error}
 \begin{tabular}{||c | c c||} 
 \hline
 RSME & Translation (m) & Rotation (deg) \\ [0.5ex] 
 \hline
 DSAC* only & 12.32 & 2.50\\ 
 Full GODSAC* & 8.99 & 2.30 \\ 
 \hline
 \end{tabular}
 \label{tab:gtsam_err}
\end{table}

\subsection{Visualization}
\begin{figure*}
    \centering
    \includegraphics[width=0.75\textwidth]{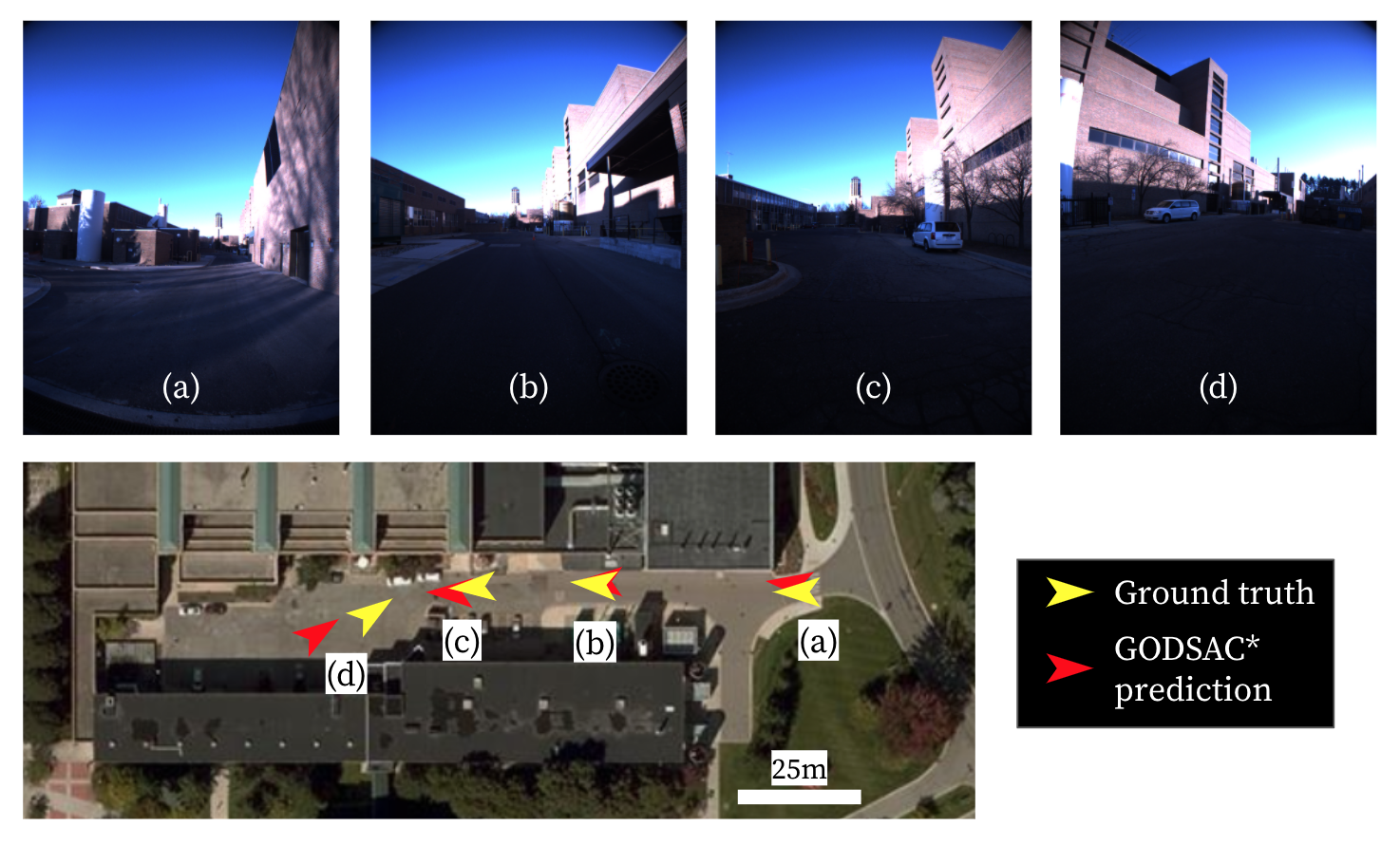}
    \caption{Visualization of SE(2) trajectory poses (bottom) and RGB images (top) of the NCLT dataset. The data is sampled approximately 25 seconds apart. Images (a), (b), and (c) are taken sequentially before a the robot turns around and captures image (d).}
    \label{fig:visualdata}
\end{figure*}

The SE(2) pose output from DSAC* was visualized through the RVIZ ROS package, and can be seen at the link provided in the abstract. A small sample of pose predictions on NCLT is also shown in Figure \ref{fig:visualdata}. This data sample was chosen because it demonstrates the pose inference from GODSAC* during a 180° turn, and the error associated with that turn. Additionally, the angular error for all points in this figure can be seen to be relatively small, yet these were all errors higher than the mean error. The top portion of the figure also shows the RGB images associated with each pose.

For a presentation of our results, see \href{https://youtu.be/3n6or2iM_vA}{https://youtu.be/3n6or2iM\_vA}

\section{Discussion}
Similar to previous sections, we analyze the results of GODSAC* after each of the two pipeline stages. 

\subsection{DSAC*}
Our results from this stage are similar to those in the DSAC* paper \cite{dsacstar}, but with the addition of NCLT. In outdoor scenes with large, generic features (e.g. trees), DSAC* does not perform very well compared with smaller scenes or scenes with unique features. For instance, unique buildings provide concrete waypoints that are learned by DSAC*. This describes why the sequence in Figure \ref{fig:visualdata} has relatively low error compared to a similar trajectory through trees. Another key example of this is the King's College dataset in Figure \ref{fig:front_king_compare}; all images of this scene were pointed at the front a large, feature-rich building. 

\subsection{GODSAC*}
The two downsides of DSAC* which we endeavored to fix in this project is the lack of support for odometry data and memorylessness (it does not leverage past predictions to improve future results). In fact, real-world ground vehicle image and odometry data can often be collected as a sequence, allowing for graph-based smoothing and mapping to form a history of past data. This is most clearly shown when comparing Figure \ref{fig:front_nclt_compare}(ii) and Figure \ref{fig:gtsam_NCLT}. The latter follows a smoother trajectory, which better matches the ground truth, which is also smooth. Thus, we are improved upon DSAC* with our implementation of in GODSAC* with iSAM2 integration, significantly lowering the prediction error.

Nevertheless, there are still outliers in the data. Whenever a trajectory starts (see right side of Figure \ref{fig:gtsam_NCLT}) or when the robot turns abruptly, the error spikes. This could prove problematic for a robot on bumpy terrain or one with jittery motion, which actually occurs in the NCLT data between $x=-120$ and $x=-200$ in the same figure. To fix this, we were exploring a few options: the use of changing pose covariances based on the confidence of the DSAC* output, the use of a deeper network with semantic information as inputs, and undistorting the RGB images (shown to improve accuracy in unpublished work). However, these were not yet completed due to time constraints.

In terms of speed, GODSAC* runs at a similar timescale to previous solutions like PoseNet++. With a single GTX 1080, GODSAC* can maintain a 100ms inference time per frame (10Hz). This is relatively good considering the high accuracy of the solution. This latency demonstrates that this solution can be used for offline pose estimation or slow-moving ground vehicles, but is not yet sufficient for real-time re-localization of fast-moving objects (e.g. a car). The latency could be improved by removing all Python and ROS dependencies (create a pure C++ package), or by exploring options like DSAC* Tiny \cite{dsacstar}.

\subsection{Future Work}
Some areas for future work include testing on more difficult datasets (e.g. poorly-lit outdoor images, dataset over larger areas, etc.), perform depth prediction on the input RGB images, or use a version of ResNet with more layers (in DSAC*).
\section{Conclusion}
We proposed GODSAC* as a camera pose estimation approach that augments pose predictions from a trained neural network with noisy odometry data through the optimization of a pose graph. GODSAC* outperforms the state-of-the-art approaches in pose estimation accuracy, both in terms of translation error and rotation error.

\bibliographystyle{plain}
\bibliography{references}
\end{document}